\documentclass[journal]{IEEEtran}
\usepackage[utf8]{inputenc}
\usepackage{graphicx}
\usepackage{amsmath,amssymb}
\usepackage{cite}
\usepackage{textcomp}
\usepackage[
    colorlinks=true,
    linkcolor=red,     % warna untuk \ref, \eqref, dll.
    citecolor=green,     % warna untuk \cite
    urlcolor=magenta    % warna untuk \url
]{hyperref}
\usepackage{caption}
\usepackage{float}
\usepackage{tabularx}
\usepackage{booktabs}
\usepackage{multirow}
\usepackage[table,xcdraw]{xcolor}
\usepackage{longtable}
\usepackage{lscape}
\usepackage{newtxtext,newtxmath}
\usepackage{arydshln}
\usepackage{fancyhdr}

\title{SKGE-SWIN: End-To-End Autonomous Vehicle Waypoint Prediction and Navigation Using Skip Stage Swin Transformer}

\author{Fachri Najm Noer Kartiman, 
        Rasim,
        Yaya Wihardi,
        Nurul Hasanah,
        Oskar Natan,
        Bambang Wahono,
        Taufik Ibnu Salim
\thanks{Fachri Najm Noer Kartiman is with Department of Computer Science, Indonesia University of Education, Bandung 40154, Indonesia (e-mail: fachrinajmnoer@upi.edu).}
\thanks{Rasim is with Department of Computer Science, Indonesia University of Education, Bandung 40154, Indonesia (e-mail: rasim@upi.edu).}
\thanks{Yaya Wihardi is with Department of Computer Science, Indonesia University of Education, Bandung 40154, Indonesia (e-mail: yayawihardi@upi.edu).}
\thanks{Nurul Hasanah is with Research Center for Smart Mechatronics, National Research and Innovation Agency, Bandung 40135, Indonesia (e-mail:nuru030@brin.go.id).}
\thanks{Oskar Natan is with the Department of Computer Science and Electronics, Gadjah Mada University, Yogyakarta 55281, Indonesia (e-mail:oskarnatan@ugm.ac.id).}
\thanks{Bambang Wahono is with Research Center for Smart Mechatronics,
National Research and Innovation Agency, Bandung 40135, Indonesia (e-mail:bamb047@brin.go.id).}
\thanks{Taufik Ibnu Salim is with Research Center for Smart Mechatronics, National Research and Innovation Agency, Bandung 40135, Indonesia (e-mail:tauf021@brin.go.id).}
}
\makeatletter
\def\ps@IEEEtitlepagestyle{%
  \fancyhf{}%
  \fancyfoot[C]{This work has been submitted to the IEEE for possible publication. Copyright may be transferred without notice, after which this version may no longer be accessible.}%
}
\makeatother
\begin{document}
\maketitle

\begin{abstract}
% Gunakan paragraf ini untuk menulis abstrak Anda.
Focusing on the development of an end-to-end autonomous vehicle model with pixel-to-pixel context awareness, this research proposes the SKGE-Swin architecture. This architecture utilizes the Swin Transformer with a skip-stage mechanism to broaden feature representation globally and at various network levels. This approach enables the model to extract information from distant pixels by leveraging the Swin Transformer's Shifted Window-based Multi-head Self-Attention (SW-MSA) mechanism and to retain critical information from the initial to the final stages of feature extraction, thereby enhancing its capability to comprehend complex patterns in the vehicle's surroundings. The model is evaluated on the CARLA platform using adversarial scenarios to simulate real-world conditions. Experimental results demonstrate that the SKGE-Swin architecture achieves a superior Driving Score compared to previous methods. Furthermore, an ablation study will be conducted to evaluate the contribution of each architectural component, including the influence of skip connections and the use of the Swin Transformer, in improving model performance.
\end{abstract}


\section{Introduction}
\label{sec:intro}
Autonomous Driving is a complex intelligent system that handles tasks ranging from perception to vehicle control, necessitating distinct modules \cite{huang2020autonomousdrivingdeeplearning}. The conventional integration of these modules, however, is often intricate and inefficient. For instance, combining information provided by perception modules—such as object detection and semantic segmentation—as input for the vehicle control module can lead to compounding errors and the loss of critical information, thus requiring numerous parameters to be tuned manually \cite{Gawlikowski2021-qx}. With the massive advancements in deep learning, many studies have shifted towards training a single model to solve tasks in an end-to-end and multitask manner \cite{chen2020interpretableendtoendurbanautonomous}\cite{bicer2019sample}. In this paradigm, the model can learn from various related tasks and deliver a final action within one integrated model. As manual tuning is no longer required, the model can autonomously leverage the features it extracts \cite{Gutierrez2020-se}.
\begin{figure}[!t]
    \centering
    \includegraphics[width=1\columnwidth]{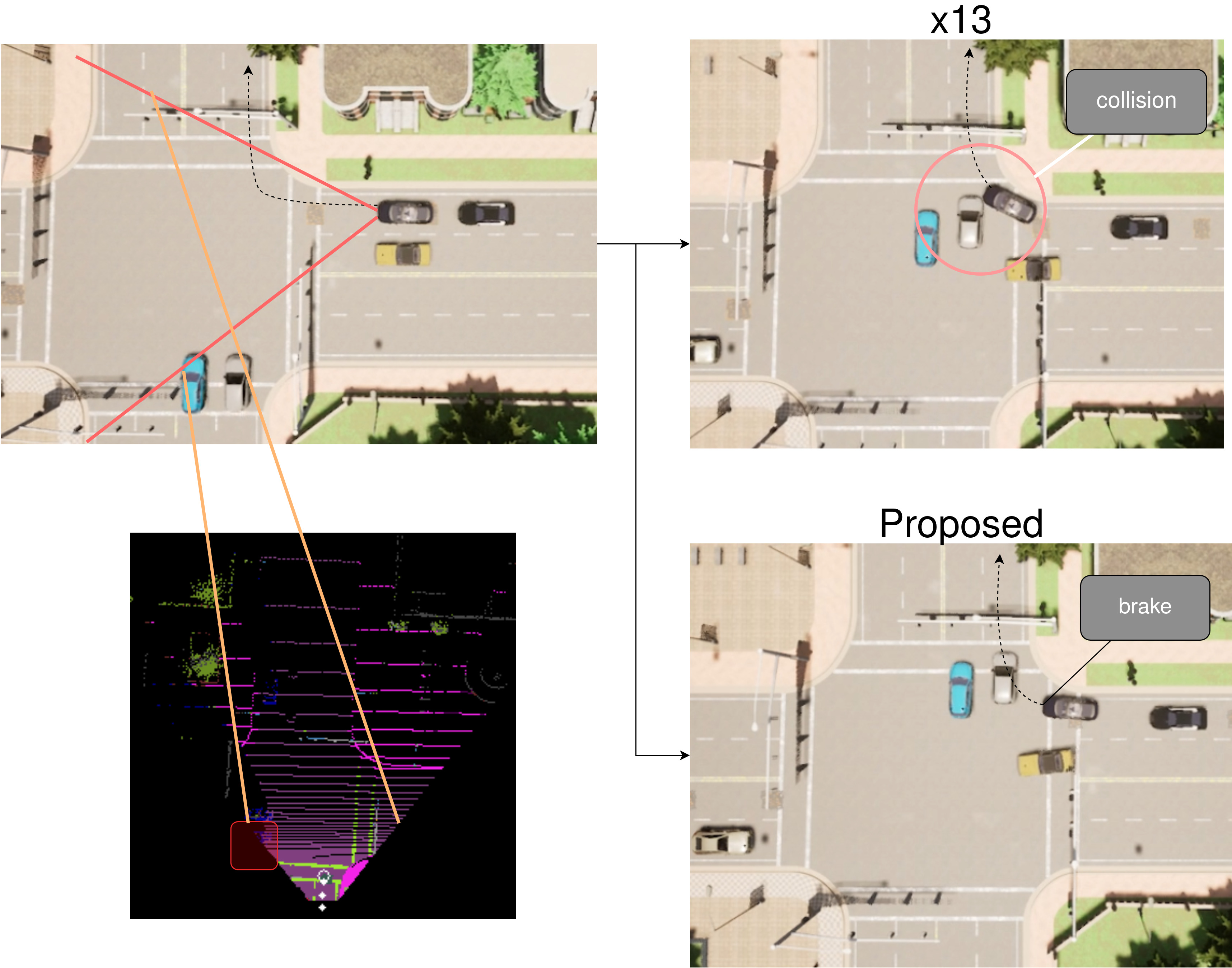}
    \caption{Pixel-to-Pixel Context Awareness for Robust Autonomous Driving.At intersections, when required to turn right or left, or when a vehicle is approaching from the opposite direction, the agent fails to perceive the vehicle's presence, leading to a risk of a side-swipe collision.}
    \label{fig:architecture-bev-tabrakan}
\end{figure}

Prior research in end-to-end autonomous driving was conducted by Natan et al. \cite{Natan2023-hi}, who utilized a Semantic Depth Cloud (SDC) to generate a bird's-eye view (BEV) by fusing RGB and depth imagery, employing a Convolutional Neural Network (CNN)-based backbone, specifically EfficientNet \cite{tan2019efficientnet}. However, this model still fails in conditions that demand global context awareness (see figure \ref{fig:architecture-bev-tabrakan}). This is because CNN-based methods are limited by their local receptive field, which makes it difficult to capture global relationships between pixels \cite{Raghu2021-jj}. Furthermore, their static convolutional weights during inference render them less adaptive to input variations. To address this, the self-attention (SA) mechanism emerged, enabling parallel processing and the learning of richer representations. This approach was subsequently adapted for computer vision tasks, for instance, in the Vision Transformer (ViT) \cite{Dosovitskiy2020-mf} and the Swin Transformer \cite{Liu2021-qz}. Among these implementations, the Swin Transformer stands out due to its Windowed Self-Attention (WSA) mechanism, which successfully reduces computational complexity from $O(N^2)$ to $O(N \log N)$ \cite{Liu2021-qz}. Additionally, the Swin Transformer features a hierarchical design that allows the model to capture both local and global information at multiple resolutions, making it highly suitable for visual perception tasks that require comprehensive contextual understanding. Nevertheless, the Swin Transformer requires information from shallow layers to reference deeper information, in order to preserve high-resolution spatial details that tend to be lost during the self-attention and patch merging processes. Therefore, skip connections become crucial for linking information from shallow to deep layers.

This research proposes the Skip Stage Swin (SKGE-Swin) model, which leverages the advantages of the Swin Transformer's hierarchical processing and self-attention mechanism, while also integrating skip connections inspired by ResNet \cite{He2015-lm} to preserve critical details from the initial to the final processing stages. This study re-engineers the encoding stage within a waypoint prediction and autonomous navigation architecture, specifically its Swin Transformer-based backbone \cite{Wu2022-oo}, to enhance the model's ability to capture global relations between pixels. This makes the model more context-aware of distant objects or conditions in the input image. The evaluation is conducted through a series of experiments to measure waypoint prediction accuracy and the model's adaptability to varying road and environmental conditions \cite{gu2019net, Xie2021-qg}. Furthermore, an ablation study will be performed to analyze the contribution of each component, including the influence of the Swin Transformer and the skip connections, on the improvement of the overall model performance.

The novelty of this research can be summarized as follows:
\begin{itemize}
\item The development of the SKGE-Swin architecture, which integrates the Swin Transformer with skip connections to enhance global contextual understanding in end-to-end autonomous driving models.
\item The application of a self-attention mechanism to the Bird's Eye View (BEV) representation to capture long-range dependencies between pixels, thereby improving waypoint prediction accuracy and vehicle navigation.
\item A performance evaluation of the SKGE-Swin model under adversarial conditions that mimic real-world scenarios, along with an analysis of the contribution of each architectural component via an ablation study.
\end{itemize}

This paper is organized as follows. In Chapter 2, we conduct a comprehensive review of related works that inspired this study. In Chapter 3, we explain the proposed model, with a particular focus on the Swin Transformer and skip connections. In Chapter 4, we describe the experiments conducted to evaluate the performance of the proposed model. In Chapter 5, we discuss the experimental results and analyze the model's performance. Finally, in Chapter 6, we conclude the study and provide suggestions for future research.
\section{Related Work}

In this section, we will discuss several related works relevant to this research topic. The primary focus is on the application of Attention mechanisms and Bird's-Eye View (BEV) in Autonomous Driving. The discussion is organized into the following subsections:

\subsection{Attention in Autonomous Driving}
Vaswani et al. \cite{Vaswani2017-wm} introduced the self-attention mechanism, which forms the basis of the Transformer architecture. This mechanism enables a model to efficiently capture dependencies between elements in data, making it highly suitable for tasks that require complex contextual understanding. The self-attention mechanism has been explored in the driving context for various tasks such as motion prediction \cite{li2020endtoendcontextualperceptionprediction} \cite{Choi2019Looking}, object detection \cite{zhao2024detrsbeatyolosrealtime}, and even end-to-end driving.

One such study by Chitta et al. \cite{Chitta2022-wl} employed an end-to-end architecture combining a CNN and a Transformer for the task of waypoint prediction in autonomous vehicles. The work by \cite{chen2017braininspiredcognitivemodel} applied self-attention to waypoint prediction in conjunction with an RNN, leveraging the time-based nature of RNNs to model self-attention for understanding historical (sequential) inputs. Similarly, Hao Shao et al. \cite{shao2022safetyenhancedautonomousdrivingusing} utilized a transformer encoder to extract features from multi-view RGB (front, right, and left) and LiDAR data, thereby broadening the model's field of view.

\subsection{BEV in Autonomous Vehicles}
\begin{figure}[!t]
    \centering
    \includegraphics[width=1\columnwidth]{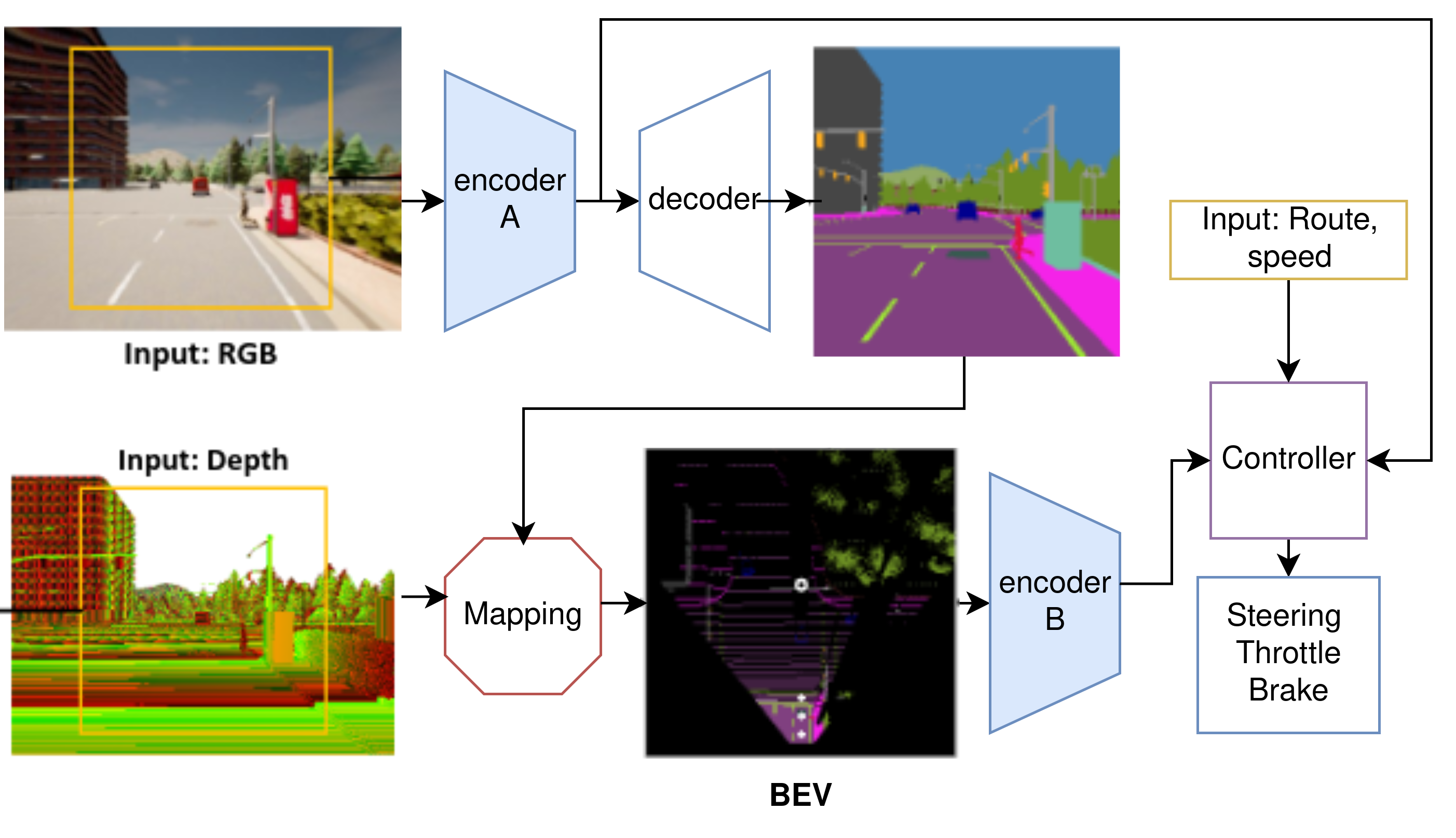}
    \caption{Overall Architecture. The blocks filled in blue indicate the parts modified from the base architecture of \cite{Natan2023-hi} to the SKGE-Swin Block. Encoder A: Semantic Segmentation Backbone, Encoder B: BEV Feature Extraction Backbone}
    \label{fig:mini-architecture}
\end{figure}

The studies by \cite{shao2022safetyenhancedautonomousdrivingusing} \cite{Chitta2022-wl} use LiDAR to obtain a Bird's-Eye View (BEV) representation of the vehicle's surrounding environment. The LiDAR representation is converted into a 2D BEV map, providing a top-down perspective that aligns with RGB inputs from cameras. Similarly, Hao Shao et al. \cite{shao2023reasonnetendtoenddrivingtemporal} utilize two BEV representations—one from LiDAR and an object density map—to accurately perceive the agent's surroundings by relying on the movement of nearby objects. However, this approach requires a separate bounding box module in real-world environments. Natan et al. \cite{Natan2023-hi} proposed an approach that combines depth mapping and semantic segmentation, termed the Semantic Depth Cloud (SDC). This method is highly effective at capturing environmental context and distinguishing nearby objects without the need for a separate, external module. Overall architecture can be seen in figure \ref{fig:mini-architecture}, sensor are used in his study are RGB and depth images, which are combined to form a Semantic Depth Cloud (SDC) representation. The representation is then transformed into a BEV format, which is used as input for the encoder. Furthermore, extracted features are processed to controller modeule, which predicts the vehicle's trajectory in the form of waypoints, steering, throttle, and brake commands. In this research study, SDC is employed as the BEV representation, from which features are extracted using the SKGE-Swin block.

\subsection{Skip Connections}
Skip connections were first introduced in the ResNet architecture \cite{He2015-lm} to address the vanishing gradient problem in very deep neural networks. This concept allows information from earlier layers to 'skip' over subsequent layers and connect directly to deeper ones, thus helping to preserve critical information and accelerate the training process. The integration of skip connections within Transformer architectures has been shown to yield specific improvements in evaluation metrics. The Swin Transformer \cite{Wu2022-oo} already incorporates skip connections within its architecture, specifically in each SW-MSA block, but not between the main stages. Natan et al. \cite{Natan2023-hi} also employed skip connections in a manner similar to their implementation in U-Net \cite{ronneberger2015unetconvolutionalnetworksbiomedical}. This research adapts the skip connection concept for the Swin Transformer to enhance the generated feature representations, allowing the model to better preserve high-resolution spatial information that supports the formation of more robust abstract spatial representations.

\section{Proposed Model}
\begin{figure*}[!t]
  \centering
  \includegraphics[width=\textwidth]{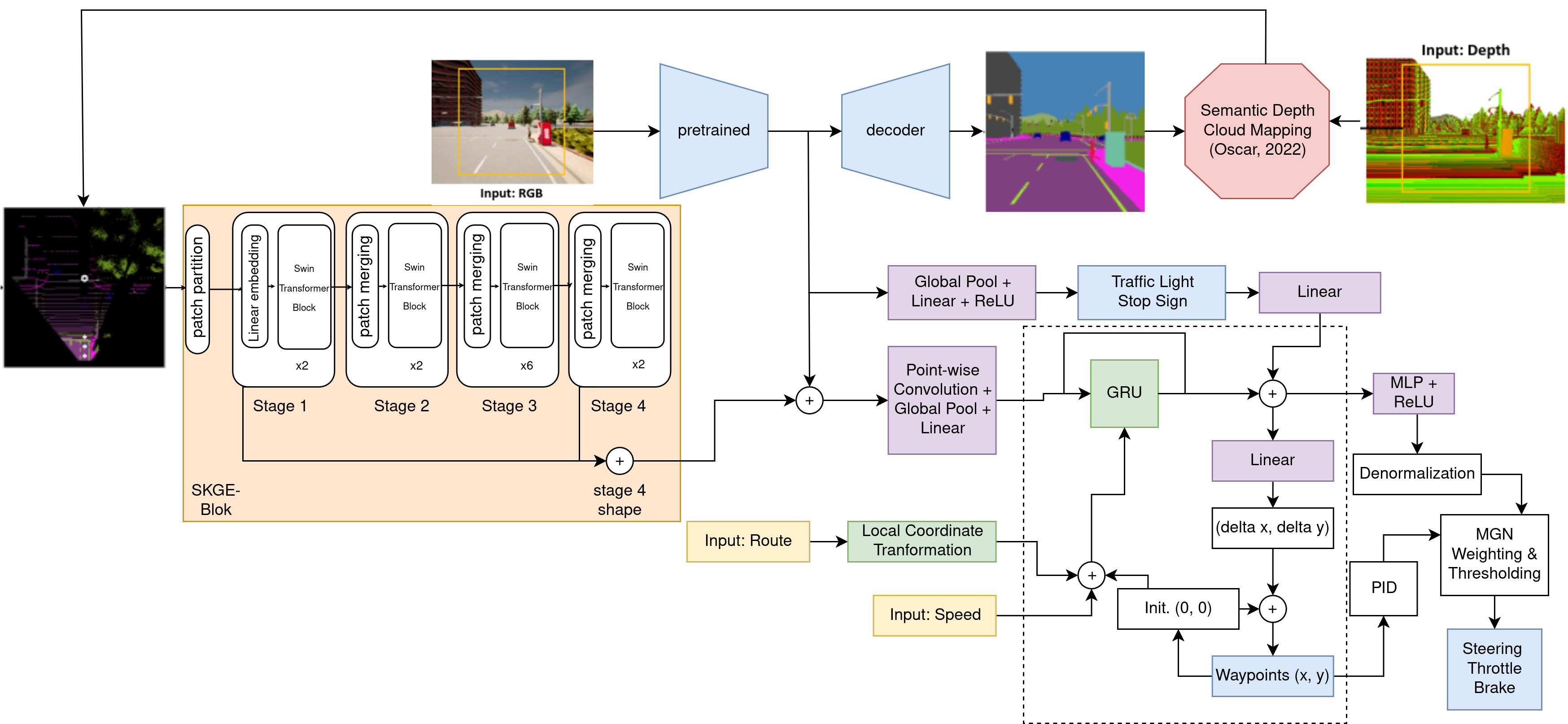}
  \caption{Proposed Model (SKGE Swin with SKGE Block). In detail view of the architecture, the SKGE Block is shown in the orange box. The SKGE Block is a Swin Transformer block with skip connections. The skip connections are applied to stages through many experiments, and the best results are obtained by applying skip connections from stage 1 to stage 4.}
  \label{fig:architecture-skge-swin-stage-14}
\end{figure*}

This section explains the architectural design of the proposed model. As shown in figure \ref{fig:architecture-skge-swin-stage-14}, this model is a development of the paper by \cite{Natan2023-hi}, utilizing the Swin Transformer with skip connections (SKGE-block). The researcher will briefly discuss the main components of this architecture. For a more in-depth explanation, please refer to \cite{Natan2023-hi}.

\subsection{Macro Design}
\label{sec:macro_design}

In this study, the macro design (see figure \ref{fig:mini-architecture}) consists of several main parts: a semantic segmentation backbone (Encoder A), a decoder, mapping, a BEV feature extraction backbone (Encoder B), and a controller. The backbone is responsible for visual feature extraction from RGB images and depth maps, the waypoint prediction serves to predict the vehicle's trajectory based on the extracted features, and the BEV Extraction is tasked with transforming SDC data into the information required for waypoint prediction. Waypoint prediction is highly related to the features generated from the preceding blocks. Therefore, the backbone and BEV Extraction also function to produce better features for waypoint prediction. The Swin Transformer \cite{Liu2021-qz} was chosen as the backbone and for BEV extraction due to its ability to capture complex spatial and temporal relationships in visual data. Furthermore, this architecture is also designed to address challenges faced in autonomous vehicle systems, such as variations in road conditions, weather, and surrounding objects. The official Pytorch and Mmsegmentation libraries were used for the implementation of the Swin Transformer, which allows for easier integration with existing deep learning frameworks.

\subsection{Skip Connection in Swin Transformer (SKGE)}

\begin{figure*}[!t]
    \centering
    \includegraphics[width=1\textwidth]{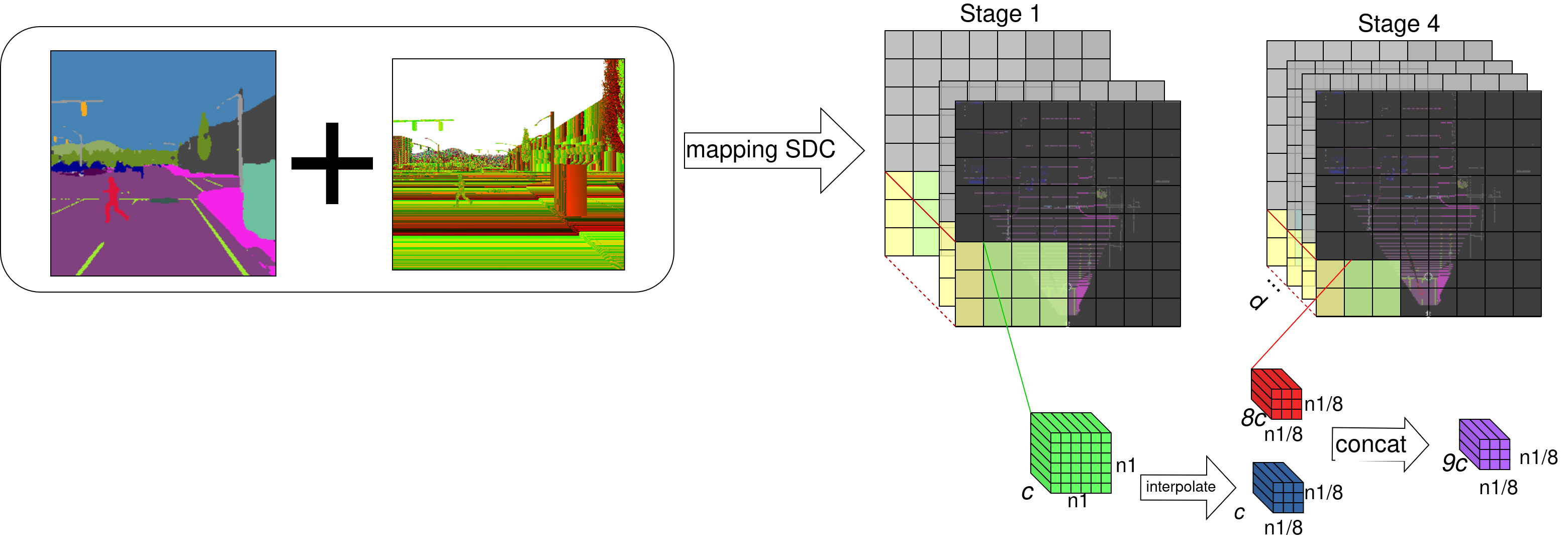}
    \caption{Skip Stage Connection in Swin Transformer.}
    \label{fig:architecture-implementasi_skge}
\end{figure*}

The Swin Transformer \cite{Liu2021-qz} is augmented with skip connections to obtain richer feature results from global features in the early stages and local features in the final stages. Experiments were conducted to test the performance of the Swin Transformer with skip connections. Skip connections are applied to every possible block of the Swin Transformer, thereby enriching the resulting feature representation. Figure \ref{fig:architecture-implementasi_skge} is an illustration of the implementation of skip connections in the Swin Transformer. In the figure, it is shown that features from stage 1 (shallow layer) are interpolated to stage 4 (deep layer) to combine information from both stages.

% Pada Gambar \ref{fig:architecture-skge-swin} diperlihatkan skip connection antar stage yang terjadi dari stage 1 dan digabung ke stage 4 menggunakan interpolasi.

% \begin{figure}[!t]
%     \centering
%     \includegraphics[width=0.4\columnwidth]{images/architecture-skge-swin.png}
%     \caption{Skip Connection pada Swin Transformer(SKGE Blok).}
%     \label{fig:architecture-skge-swin}
% \end{figure}

\begin{equation}
    \label{eq:bilinear_interpolation}
      \begin{aligned}
      f(x, y) \approx \frac{1}{(x_2 - x_1)(y_2 - y_1)} \Big[ &
      Q_{11}(x_2 - x)(y_2 - y) + \\
      & Q_{21}(x - x_1)(y_2 - y) + \\ 
      & Q_{12}(x_2 - x)(y - y_1) + \\
      & Q_{22}(x - x_1)(y - y_1)
      \Big]
      \end{aligned}
    \end{equation}
Bilinear interpolation, equation \ref{eq:bilinear_interpolation}, is used to estimate the value at a point based on the values of its four nearest neighboring points. This method provides smoother results compared to nearest-neighbor interpolation, as it considers value changes in two directions (horizontal and vertical). After interpolation, the two features are combined by addition, resulting in richer features from the sum of the desired stages.

\subsection{Swin Transformer Varians}
% The Swin Transformer has several different variants, each with different characteristics and sizes. Table \ref{tab:swin_variants} shows a comparison between these variants, including patch size, embedding dimension, number of layers, number of heads, number of parameters (M), and FLOPs (G). Each variant has its own advantages and disadvantages, so the selection of the right variant is crucial for achieving optimal performance in specific tasks.

% \begin{table*}[!t]
%     \centering
%     \caption{Swin Transformer Varians}\label{tab:swin_variants}
%     \resizebox{\textwidth}{!}{%
%     \begin{tabular}{lcccccc}
%     \hline
%     \textbf{Variant} & \textbf{Patch Size} & \textbf{Embed Dim} & \textbf{Layers} & \textbf{Heads} & \textbf{Params (M)} & \textbf{FLOPs (G) @224} \\
%     \hline
%     Swin-T (Tiny)  & 4×4 & 96  & [2, 2, 6, 2]  & [3, 6, 12, 24]  & 28M  & 4.5   \\
%     Swin-S (Small) & 4×4 & 96  & [2, 2, 18, 2] & [3, 6, 12, 24]  & 50M  & 8.7   \\
%     Swin-B (Base)  & 4×4 & 128 & [2, 2, 18, 2] & [4, 8, 16, 32]  & 88M  & 15.4  \\
%     Swin-L (Large) & 4×4 & 192 & [2, 2, 18, 2] & [6, 12, 24, 48] & 197M & 34.5  \\
%     \hline
%     \end{tabular}%
%     }
%     \end{table*}
    
In this experiment, the Swin Transformers used were Swin-B (Base) and Swin-T (Tiny) from \cite{Liu2021-qz}, which have been modified by adding skip connections. Swin-B (Base) has a larger size and more parameters compared to Swin-T (Tiny), so it can provide better performance in complex tasks. However, Swin-T (Tiny) also has an advantage in terms of computational and memory efficiency, making it suitable for use on devices with limited resources. Swin-T (Tiny) gains an advantage when used on limited datasets, whereas Swin-B (Base) is superior on larger and more complex datasets.

\section{Experiment Setup}

\subsection{Dataset and Representation}
In this research stage, the data that was used as input for training and testing the autonomous vehicle model was collected from various sensors commonly used in such systems. The obtained data provided a comprehensive understanding of the vehicle's surrounding environmental conditions and the state of the vehicle itself, which was crucial in supporting visual perception, navigation, and decision-making tasks in autonomous vehicle systems. The following sub-sections provided more detail about the collected data.

In this study, the approach used was imitation learning, specifically behavior cloning. The goal of this approach was to learn a policy $\pi$ by imitating the behavior of an expert who had a policy $\pi^*$. The policy was defined as a mapping function from input to trajectory (waypoints), steering, acceleration (throttle), and brake, which could be approached with the supervised learning paradigm. The simulator used was CARLA version 0.9.10.1 to generate the dataset used in training and validating the model. All of CARLA's built-in maps and weather conditions were used to create a varied simulation environment. Additionally, non-player characters (NPCs) such as vehicles and pedestrians were also simulated to mimic real-world conditions. The ego vehicle was controlled by a privileged agent, which recorded a comprehensive set of data every 500 ms. Each data point consisted of the primary visual inputs from a forward-facing camera: an RGB image and its corresponding depth map. Accompanying the image was a pixel-wise semantic segmentation map, categorized into 23 classes as detailed in Table \ref{tab:carla_dataset_setting}. The vehicle's own state was captured through its speed (in m/s) and steering angle, while the control commands from the agent were logged as the applied throttle and brake values. For navigational context, each data point also contained the expert's trajectory, defined as a sequence of waypoints, and the environmental state, which included the status of nearby traffic lights and the detection of any stop signs in the vehicle's path.

The trajectory was defined in the form of 2D waypoints in a bird's eye view (BEV) space, in the vehicle's local coordinates. Vehicle controls were recorded as steering, throttle, and brake values at the time of data capture. For comparison with other models, point cloud data from LiDAR was also collected. The route configuration followed \cite{prakash2021multi}, where the expert was given a number of predefined routes with a sequence of GPS coordinates. 

There were three categories of routes: Long routes: 1000–2000 meters, ±10 intersections per route, Short routes: 100–500 meters, ±3 intersections per route, Mini routes: one turn or one straight path. The number of routes per category depended on the map's characteristics. Town01 to Town06 had all three types of routes, whereas Town07 and Town10 only had short and mini routes. The type of dataset generated was for evaluation in clear noon weather only. This dataset was denoted as equation \ref{eq:dataset_eq}:
\begin{equation}
  \label{eq:dataset_eq}
    D = \{(X_i, Y_i)\}_{i=1}^{J}
\end{equation}
where J was the number of data points. Input ($X$) included RGB Image, Depth map, Speed, Routepoints. Output ($Y$) included Semantic segmentation (ground truth), Waypoints, Vehicle controls (steering, throttle, brake), Traffic light status, Stop sign status

Although some inputs like point clouds and navigation commands were used in the dataset, the developed model did not depend on them. The following was a more detailed explanation of the data representation used in this research.
RGB Image and Depth Map were obtained at a resolution of 300×400 pixels, then cropped to 256×256 pixels. The representation of RGB images and depth was $R \in {0, ..., 255}^{3 \times 256 \times 256}$. LiDAR Point Clouds were represented as a 3D tensor with dimensions of$\mathbb{R}^{4 \times 13716}$ where 4 was the number of features (x, y, z, intensity) and 13716 was the number of points captured from the LiDAR. This point cloud data provided information about the position and intensity of light reflected by objects around the vehicle. Semantic Segmentation was represented as a binary tensor $R \in \{0, 1\}^{23 \times 256 \times 256}$, where 23 was the number of object classes as shown in table \ref{tab:carla_dataset_setting}. Each pixel in the semantic segmentation image was labeled according to the detected object class, allowing the model to understand the visual context around the vehicle. Waypoints were denoted as $\omega_i^\rho = (x_i, y_i)$ for $i = 1, 2, 3$ based on the desired number of waypoint predictions, which was three. The center point of the BEV coordinates (0, 0) was at the bottom center of the ego vehicle. Vehicle Control\label{sec:kontrol_kendaraan} (steering, throttle, brake) were normalized to an appropriate range, then de-normalized to their original range: Steering $\in [-1, 1]$, Throttle $\in [0, 0.75]$, Brake $\in [0, 1]$, Traffic light status and stop signs were labeled 1 if they appeared, 0 otherwise. Speed, measured in meters per second (m/s), and GPS location were recorded sparsely to provide more detailed information about the vehicle's movement.

The dataset used in this research was obtained from the CARLA simulation platform, which provided various realistic scenarios with different weather conditions, times of day, and road types. The use of a simulation dataset offered the advantage of testing the performance of autonomous driving models without needing a physical vehicle, allowing for the simulation of various conditions faced in the real world. By utilizing this simulation data, the research could obtain a rich and comprehensive representation for training and testing the model under various environmental conditions. The data used in this study referred to an existing dataset from previous research \cite{Chitta2022-wl}, which included various scenarios and configurations. Table \ref{tab:carla_dataset_setting} further explains the data collection settings on the CARLA platform.

\begin{table}[!t]
\centering
\caption{Carla Dataset Setting}
\label{tab:carla_dataset_setting}
\begin{tabularx}{\columnwidth}{lX}
\hline
\textbf{Maps} & Town01, Town02, Town03, Town04, Town05, Town06, Town07, Town10 \\
\hline
\textbf{Route sets*} & 
Long (1000--200m), Short (100--500m), Tiny (one turn or one go-straight) \\
\hline
\textbf{Weather presets} & ClearNoon \\
\hline
\textbf{Non-playable characters} & Vehicles (truck, car, bicycle, motorbike) and pedestrians \\
\hline
\textbf{Object classes} & 
0: Unlabeled, 1: Building, 2: Fence, 3: Other, 4: Pedestrian, 5: Pole, 6: Road lane, 7: Road, 8: Sidewalk, 9: Vegetation, 10: Other vehicles, 11: Wall, 12: Traffic sign, 13: Sky, 14: Ground, 15: Bridge, 16: Rail track, 17: Guard Trail, 18: Traffic light, 19: Static Object, 20: Dynamic Object, 21: Water, 22: Terrain \\
\hline
\textbf{CARLA version} & 0.9.10.1 \\
\hline
\end{tabularx}
\end{table}

The cleaning performed on this dataset included the removal of incomplete or irrelevant data, such as data missing RGB images, depth maps, or semantic segmentation. Additionally, data with extreme values or outliers in Driving Score and Infraction Penalty were also removed to ensure that the trained model was not affected by unrepresentative data. RGB, Depth, and Segmentation images were obtained at a resolution of 300×400 pixels, then cropped to 256×256 pixels. The representation for RGB and depth images was $R \in {0, ..., 255}^{3 \times 256 \times 256}$. The actual depth ($R^{{dec}}_i$) for each pixel was calculated using the following equation: The collected depth maps, in RGB format, were decoded to obtain the actual depth. Each pixel in the depth map was processed using a formula to calculate the depth in meters based on the RGB color values in the image.
The formula used to decode the depth value was in equation \ref{eq:depth_decoding} below \cite{carladepthcamera}:
\begin{equation}
\label{eq:depth_decoding}
\mathbb{R}_i^{\text{dec}} = \frac{R_i + 256 G_i + 256^2 B_i}{256^3 - 1} \times 1000,
\end{equation}

With $R_i, G_i, B_i$ being the 8-bit pixel values, and 1000 being the camera's depth range in meters. In the first iteration of the loop, the GRU used the features as the \textit{initial hidden state} and received three types of input: the current \textit{waypoint} coordinates in BEV space (local vehicle coordinates), the route location coordinates transformed into BEV space, and the current vehicle speed (in m/s). It should be noted that the initial value for the \textit{waypoint} coordinates was the local vehicle coordinate, which was always at the point $(0, 0)$ and located at the bottom-center of the \textit{semantic depth cloud} (SDC). The transformation from global coordinates $(x_g, y_g)$ to local coordinates $(x_l, y_l)$ could be done with equation~\eqref{eq:transform}:

\begin{equation}
\begin{bmatrix}
x_l \\
y_l
\end{bmatrix}
=
\begin{bmatrix}
\cos(90 + \theta_v) & -\sin(90 + \theta_v) \\
\sin(90 + \theta_v) & \cos(90 + \theta_v)
\end{bmatrix}^T
\begin{bmatrix}
x_g - x_{vg} \\
y_g - y_{vg}
\end{bmatrix}
\label{eq:transform}
\end{equation}
The relative distance was obtained by subtracting the vehicle's global coordinates from the route's global coordinates. The rotation angle $90 + \theta_v$ was used in the rotation matrix because the GPS compass points north. Then, the next \textit{hidden state} generated from the GRU was \textit{biased} by the predicted traffic light status and the presence of a stop sign, which had been previously encoded using a \textit{linear layer} and added through element-wise summation. For \textit{waypoint} prediction, the \textit{biased hidden state} was then translated into $\Delta x$ and $\Delta y$ values through a \textit{linear layer}. The next \textit{waypoint} was calculated with equation \ref{eq:waypoint}:
    
\begin{equation}
\label{eq:waypoint}
x_{i+1}, y_{i+1} = (x_i + \Delta x), (y_i + \Delta y)
\end{equation}
    
where $i$ was the step index in the loop. In the next iteration, the GRU used the current \textit{hidden state} (before being \textit{biased} by the traffic light and stop sign status) to predict a new \textit{hidden state}, with the first \textit{waypoint} replacing the vehicle's local coordinate point $(0,0)$ as input, along with the same route location and speed data. After the loop was completed, three predicted \textit{waypoints} and one latent representation were obtained. Data point clouds from LiDAR were converted into a 2-bin histogram in a 2D BEV image of size $\mathbb{R}^{2 \times 256 \times 256}$, representing points above and below the ground plane. The data split was done by dividing the dataset into three parts: training data, validation data, and testing data. Town01, Town02, Town03, Town04, Town06, Town07, and Town10 were used as training data. Town05 was used as testing data. This division was done to ensure that the model could learn from various different conditions and scenarios, and to test the model's performance on data it had never seen before.

\subsection{Training Configuration}
\label{sec:training_configuration}
This stage aimed to evaluate the results of each model test conducted previously. The multi-task learning approach requires several loss functions to train the model effectively:

\textbf{Semantic Segmentation Loss:} Combined binary cross-entropy and Dice loss, which could be calculated by equation \ref{eq:loss_segmentasi_semantik}. This combination offered the advantages of both distribution-based and region-based approaches, thereby improving segmentation accuracy, especially for imbalanced classes and objects of varying sizes.
\begin{equation}
\label{eq:loss_segmentasi_semantik}
\begin{aligned}
\mathcal{L}_{\text{SEG}} = 
\left( \frac{1}{N} \sum_{i=1}^{N} y_i \log(\hat{y}_i) + (1 - y_i) \log(1 - \hat{y}_i) \right) \\ + \left( 1 - \frac{2 |\hat{y} \cap y|}{|\hat{y}| + |y|} \right)
\end{aligned}
\end{equation}
where $y$ was the ground truth label, $\hat{y}$ was the model's prediction, and $N$ was the number of pixel elements in the output layer of the semantic segmentation decoder.
        
% \textbf{Loss untuk Tugas Lain:} Loss L1 digunakan untuk tugas-tugas lain, termasuk Loss status lampu lalu lintas, tanda berhenti, kemudi, gas, rem, jalur waypoint.
% $y_i$ dan $\hat{y}i$ merupakan nilai elemen ke-$i$ dari ground truth $y$ dan prediksi $\hat{y}$ secara berurutan. Sementara itu, untuk tugas-tugas lainnya seperti deteksi status lampu lalu lintas ($\mathcal{L}_{\text{TL}}$), deteksi rambu berhenti ($\mathcal{L}_{\text{SS}}$), kemudi ($\mathcal{L}_{\text{ST}}$), throttle ($\mathcal{L}_{\text{TH}}$), rem ($\mathcal{L}_{\text{BR}}$), dan prediksi waypoints ($\mathcal{L}_{\text{WP}}$), digunakan fungsi kehilangan L1 loss sederhana sebagaimana dirumuskan dalam Persamaan \ref{eq:loss_l1}
\textbf{Loss for Other Tasks:} L1 loss was used for other tasks, including the loss for traffic light status, stop signs, steering, throttle, brake, and waypoint trajectory.
$y_i$ and $\hat{y}i$ were the values of the $i$-th element of the ground truth $y$ and the prediction $\hat{y}$ respectively. Meanwhile, for other tasks such as traffic light status detection ($\mathcal{L}_{\text{TL}}$), stop sign detection ($\mathcal{L}_{\text{SS}}$), steering ($\mathcal{L}_{\text{ST}}$), throttle ($\mathcal{L}_{\text{TH}}$), brake ($\mathcal{L}_{\text{BR}}$), and waypoint prediction ($\mathcal{L}_{\text{WP}}$), a simple L1 loss function was used as formulated in equation \ref{eq:loss_l1}

\begin{equation}
    \label{eq:loss_l1}
    \mathcal{L}_{\{\text{TL}, \text{SS}, \text{ST}, \text{TH}, \text{BR}, \text{WP}\}} = |\hat{y} - y|
    \end{equation}
    
It should be noted that only $\mathcal{L}_{\text{WP}}$ needed to be averaged because there were three predicted waypoints. The model did not directly predict the x, y coordinates, but rather predicted their differences  ($\Delta x$ and $\Delta y$) in meters. Therefore, the waypoint values had to be first calculated using equation \ref{eq:waypoint} before the loss calculation. Meanwhile, the predictions for vehicle controls like steering, throttle, and brake had to be denormalized as explained in \ref{sec:kontrol_kendaraan}. Finally, the total loss function covering all tasks could be calculated using equation \ref{eq:loss_total}.
\begin{equation}
    \label{eq:loss_total}
    \begin{aligned}
    \mathcal{L}_{\text{TOTAL}} = \alpha_1 \mathcal{L}_{\text{SEG}} + \alpha_2 \mathcal{L}_{\text{TL}} + \alpha_3 \mathcal{L}_{\text{SS}} + \alpha_4 \mathcal{L}_{\text{ST}} + \\ \alpha_5 \mathcal{L}_{\text{TH}} + \alpha_6 \mathcal{L}_{\text{BR}} + \alpha_7 \mathcal{L}_{\text{WP}}
    \end{aligned}
\end{equation}
The loss weights for each task were denoted as $\alpha_1$ to $\alpha_7$. To adaptively adjust these weights in each training epoch, we used an adaptive loss weight adjustment algorithm called Modified Gradient Normalization (MGN) \cite{natan2022towards}.

In a multi-task model, it was important to balance learning between tasks by modifying the gradient signals so that the model did not focus too much on any single task. The model was trained using the Adam optimizer with a decoupled weight decay of 0.001 \cite{loshchilov2017decoupled}. The initial learning rate was set to 0.0001 and was halved if there was no decrease in the validation loss for three consecutive epochs. The training process was automatically stopped if there was no improvement in 15 consecutive epochs to avoid wasting unnecessary computational resources. This model was implemented using the mmsegmentation \cite{mmseg2020} framework and Official PyTorch \cite{paszke2019pytorch} with a batch size of 8.

\subsection{Task Evaluation}
\label{sec:task_evaluation}
Evaluation metrics were used to assess the balance of the model's performance across various tasks. The evaluation was conducted on the testing data (Town 05 long route).
    
For the semantic segmentation metric, Intersection over Union (IoU) (see equation \ref{eq:iou_segmentasi_semantik}) was used, calculated with the following formula:
\begin{equation}
    \label{eq:iou_segmentasi_semantik}
    IoU_{\text{SEG}} = \frac{|\hat{y} \cap y|}{|\hat{y} \cup y|}
\end{equation}
with $y$ being the ground truth label, $\hat{y}$ the model's prediction. For the traffic light and stop sign metrics, simple accuracy (see equation \ref{eq:acc_lampu_rambu}) was used, calculated with the following formula:
\begin{equation}
    \label{eq:acc_lampu_rambu}
    Acc_{\text{(TL,SS)}} = \frac{TP + TN}{TP + TN + FP + FN}
\end{equation}
with TP being true positive, TN true negative, FP false positive, and FN false negative. This metric measured how well the model classified the status of traffic lights and stop signs. And for the waypoints and vehicle control (steering, throttle, brake) metrics, Mean Absolute Error (MAE) was used, same as the L1 loss formula in subsection \ref{sec:training_configuration}.

\subsection{Evaluation Metrics}
In the process of evaluating the performance of the autonomous driving model, a number of metrics were used to measure the model's capabilities in various aspects of driving. This study referred to the CARLA leaderboard evaluation setup and used the Driving Score (DS) as the main metric. The higher the DS value, the better the model's performance. Route Completion (\(RC_i\))(see equation \ref{eq:route_completion}) value was obtained by dividing the route distance successfully traveled by the vehicle on route (i) by the total length of that route.
\begin{equation}
\label{eq:route_completion}
RC_i = \frac{\text{Distance completed}}{\text{Total route length}}
\end{equation}
However, if the vehicle went off-road, such as driving over a sidewalk, the part of the path traveled while off-road was not counted, thus reducing the \(RC_i\) value. The infraction penalty (\(IP_i\))(see equation \ref{eq:infraction_penalty}) was obtained by calculating the product of the penalty weights for each type of infraction raised by the vehicle during the route.
\begin{equation}
\label{eq:infraction_penalty}
    IP_i = \prod_{j=1}^{M} \left( p_i^j \right)^{\#\text{infractions}_j},
\end{equation}
Where \(p_i^j\) was the penalty weight for the \(j\)-th type of infraction, \(\#\text{infractions}_j\) was the number of infractions of that type, and $M$ was the number of penalty types in Carla. The best score for the infraction penalty was 1, which meant no infractions occurred. If the vehicle committed an infraction, the penalty would reduce the \(IP_i\) value according to the type and number of infractions that occurred. Ordered by the severity of their impact, the infraction types $M$ and penalties \(p_i^j\) were as follows:
\begin{itemize}
    \item Collision with pedestrian: 0.50
    \item Collision with vehicle: 0.60
    \item Collision with static object: 0.65
    \item Red light violation: 0.70
    \item Stop sign violation: 0.80
\end{itemize}
  
\begin{equation}
\label{eq:driving_score}
    DS = \frac{1}{N_r} \sum_{i=1}^{N_r} RC_i IP_i
\end{equation}
  The Driving Score (DSi)(see equation \ref{eq:driving_score}) for route \(i\) was a performance indicator obtained by considering several main components. These components included the multiplication of the route completion percentage for route \(i\) (Route Completion or \(RC_i\)), which reflected how much of the route was successfully completed, and the Infraction Penalty \(IP_i\) which recorded various infractions that occurred during the journey. This assessment was performed by integrating the evaluation results from the total number of evaluated routes (\(N_r\)) thus providing a comprehensive picture of the performance in completing all specified routes.
\begin{table}[!t]
    \centering
    \caption{Comparative Quantitative Result With Prior Model}
    \label{tab:perbandingan_ds}
    \begin{tabular}{p{2.5cm}lcccc}
    \toprule
    \textbf{Model} &\textbf{Quantization}& \textbf{DS↑} & \textbf{RC↑} & \textbf{IP↑} \\
    \midrule
    x13 & float 32& 29.7100 & \textbf{86.8683} & 0.3421 \\
    \hline
    % *SKGE-Swin-base (stage 2→3) & 24.6327 & 68.9117 & 0.3604 \\ 
    % *SKGE-Swin-base (stage 1→4) + lidar & 31.5026 & 74.2968 & 0.4245 \\ 
    % **SKGE-Swin-tiny (stage 4) & 32.7298 & 79.2671 & 0.4106 \\ 
    SKGE-Swin-tiny (stage 1→4) \textbf{(Proposed)} &float 16 & 31.5970 & 62.7036 & \textbf{0.5040} \\ 
    SKGE-Swin-tiny (stage 1→4) \textbf{(Proposed)} &float 32& \textbf{37.1046} & 82.8075 & 0.4491 \\
    \hdashline
    Expert  &float 32& 42.6343 & 85.5922 & 0.4981  \\
    \bottomrule
\end{tabular}
\end{table}

\section{Result and Discussion}

\subsection{Comparison of Driving Score in CARLA Simulation}

In this section, we compared the performance of various models used for the end-to-end autonomous driving task based on three main metrics: Route Completion (RC), Infraction Penalty (IP), and Driving Score (DS). The assessment was conducted in the CARLA simulation, which allows for the testing of various architectural approaches and input/output configurations in a virtual driving scenario that resembled real-world conditions. Table \ref{tab:perbandingan_ds} presents a comprehensive comparison of all tested models. In general, the SKGE-Swin-tiny (stage 1→4) model from the Official PyTorch library ranked at the top with a DS of 37.10, approaching human driver performance. The SKGE-Swin-base (stage 1→4) + LiDAR model showed competitive performance with a DS of 31.50.

\subsection{Model Performance on FPS and Memory Usage}
In the process of evaluating the performance of the models used in this research, two main metrics of primary concern were Frames Per Second (FPS) and GPU memory usage (VRAM Usage). FPS describes how many frames the model can process in one second and serves as a key indicator for assessing the model's suitability for use in real-time systems, especially in applications like autonomous vehicles. The higher the FPS value, the faster the model can respond to visual inputs directly, thus reducing latency and improving the system's safety or comfort. On the other hand, GPU memory usage reflects the model's efficiency in utilizing hardware resources. In embedded system environments like NVIDIA Jetson or other edge computing systems, the available VRAM capacity is very limited, so excessive memory usage can cause the system to run out of resources, leading to performance degradation, or even execution failure.

One of the main obstacles in implementing Transformer-based architectures, especially in the domain of visual perception for autonomous vehicles, was the significantly higher memory requirement compared to convolutional architectures (CNN). As seen in table \ref{tab:fps-memory-model}, Swin Transformer models consumed nearly twice as much VRAM as lightweight models like x13 or other CNN variants. This was consistent with findings in the literature, as described by \cite{Dosovitskiy2020-mf} in the Vision Transformer (ViT), that the self-attention mechanism requires quadratic computational complexity with respect to the number of input tokens, causing the memory load to increase drastically when applied to high-resolution images. Although the Swin Transformer adopts a local-window attention approach to suppress complexity, the overhead in parameters and memory was still unavoidable.

Nevertheless, the trade-off between memory consumption and inference speed was not linear. For example, the implementation of SKGE-Swin-tiny (1→4) with float16 showed a significant FPS increase (27.49 FPS) compared to its float32 version (22.82 FPS), even though their memory usage was identical. This indicates that inference optimizations like mixed-precision (float16) are very useful for deployment on edge devices like NVIDIA Jetson or other embedded GPUs that have Tensor Core support and can significantly accelerate performance without sacrificing accuracy \cite{micikevicius2018mixedprecisiontraining}. This provides strong evidence that the choice of numerical representation can be key in accelerating inference, even more important than simply choosing the lightest model.

On the other hand, CNN models like x13 and xconvnext remained important baselines due to their memory efficiency. For example, x13 only used 556 MiB and still achieved an FPS of 23.85, not far from Transformers that required more than 1000 MiB for comparable results. According to \cite{han2016deepcompressioncompressingdeep} in the discussion of deep compression and model efficiency, architectures that rely on local convolution like CNNs tend to be easier to optimize with techniques like pruning and quantization without drastically sacrificing accuracy. Therefore, in scenarios of embedded systems with limited power and memory, CNNs still held a competitive advantage in terms of efficiency.

However, there was an important context in model selection. Transformers, particularly the Swin architecture which has a hierarchical structure and integrates global attention, theoretically have richer spatial and semantic representation capabilities \cite{Liu2021-qz}. This ability is crucial when the input contains complex information or poor lighting conditions, which often occur in highway environments. Therefore, in autonomy applications that emphasize robustness over mere efficiency, Transformers are still worth considering despite being more computationally expensive. Moreover, models like the cognitive transfuser are capable of combining various modalities (RGB, lidar, BEV) and still maintaining competitive performance.
\begin{table}[!t]
    \centering
    \caption{FPS and Model Memory Usage}
    \label{tab:fps-memory-model}
    \begin{tabular}{p{4cm}cc}
    \toprule
    \textbf{Model Name} & \textbf{VRAM Usage(MiB)↓} & \textbf{FPS↑} \\
    \midrule
    x13 & \textbf{556} & 23.8599 \\
    \hline
    SKGE-Swin-base (stage 4→1)* & 1402 & 21.6371 \\
    SKGE-Swin-base (stage 1→4)* & 1370 & 20.9001 \\
    SKGE-Swin-base (stage 1,2,3→4)* & 1382 & 20.3343 \\
    SKGE-Swin-base (stage 3)* & 1240 & 20.7033 \\
    SKGE-Swin-base (stage 2→3)* & 1370 & 20.2175 \\
    \hline
    SKGE-Swin-base (stage 3)* & 1282 & 22.6275 \\
    SKGE-Swin-base (stage 2→4)* & 1282 & 22.6029 \\
    SKGE-Swin-base (stage 1→4)* & 1282 & 22.7276 \\
    SKGE-Swin-base (stage 1→4) + lidar* & 1410 & 22.7458 \\
    SKGE-Swin-tiny (stage 4)* & 1176 & 23.3165 \\
    SKGE-Swin-tiny (stage 4)** & 1034 & 21.3637 \\
    SKGE-Swin-tiny (stage 1→4)** & 1048 & 22.8296 \\
    SKGE-Swin-tiny (stage 4) float16** & 1016 & 23.6429 \\
    SKGE-Swin-tiny (stage 1→4) float16** & 1016 & \textbf{27.4900} \\
    \bottomrule
    \end{tabular}
    \vspace{2mm}
    
    \raggedright
    {\footnotesize
    * Using mmsegmentation implementation. \\
    ** Using Official PyTorch implementation.
    }
\end{table}

\subsection{Ablation Study}
The architectural transformation from CNN to Transformer not only provides advantages in global representation but also opens up opportunities for more flexible integration between network modules. In this context, we further examined the comparison between CNNs and Transformers in semantic segmentation and BEV extraction tasks, as well as how approaches like skip stage connection in the Swin Transformer could be utilized to overcome the limitations of hierarchical architectures. In the model performance testing conducted in this study, the evaluation metric was the Test Loss as is explained in section \ref{sec:task_evaluation}, where a lower value indicates better model performance. The evaluation was carried out on various architectural approaches and feature extractions with the aim of finding the most optimal configuration for understanding visual context in a BEV (Bird’s Eye View) based driving environment, as well as the fusion of multimodal data such as LiDAR.
\subsubsection{SKGE-swin on Semantic Segmentation Task and Skip Stage Connection in Swin Transformer}

In the SKGE-Swin-base (stage 4→1) design, the results were unsatisfactory, see table \ref{tab:perbandingan_test_loss_backbone}, which obtained results far above the average test loss of other models and was supported by table \ref{tab:detail_loss_backbone_semantic} with a high brk\_metric result, although the DS remained close (see table \ref{tab:perbandingan_ds_backbone}). This was because it used a revert skip stage, which could cause feature mismatch. \cite{zeiler2013visualizingunderstandingconvolutionalnetworks} supports the statement that features taken from deeper layers to earlier layers will cause a semantic mismatch. The proper use of skip connections is important for referencing deeper information to maintain high-resolution spatial information that tends to be lost during the self-attention and patch merging processes in the Swin transformer. Therefore, the skip stage connection widely used in the experiments was a skip stage from an earlier layer to a deeper layer.

In the Swin Transformer research \cite{Liu2021-qz}, it is explained that its model learns in a manner very similar to CNNs. Therefore, the researcher used the reference from \cite{zeiler2013visualizingunderstandingconvolutionalnetworks} which states that in the early stages of training, the model learns basic features like edges, corners, and textures, while in the final stages, it learns more complex features like shapes and objects. Consequently, the researcher conducted experiments using different stage extraction configurations, such as SKGE-Swin-base (stage 1→4), SKGE-Swin-base (stage 2→3), SKGE-Swin-base (stage 1,2,3→4), and SKGE-Swin-base (stage 3) to see how these changes affected model performance in the semantic segmentation task.

The experimental results showed that the RGB backbone model that achieved the best result on the test metric was the SKGE-Swin-base (stage 1→4) model (see table \ref{tab:perbandingan_test_loss_backbone}), which was better than other semantic segmentation backbone models. However, these models that replaced the RGB backbone in the semantic segmentation task were still unable to surpass the baseline x13 model, which used a CNN as its backbone. This is due to locality, where adjacent pixels tend to be interrelated. Convolution operations with small kernels (e.g., 3x3) are inherently designed to capture these local patterns (edges, corners, textures). In a Spatial Hierarchy, simple features can be combined to form more complex features, and the layers of convolution and pooling in a CNN naturally create this hierarchical and multi-scale feature representation. For semantic segmentation, whose goal is to label every pixel, both of these biases are crucial. The model must understand the local context very well to determine the precise boundaries of objects.

\subsubsection{SKGE-swin on BEV Feature Extraction Task}

From the previous explanation, the researcher noted that the SKGE-Swin model, which used the Swin Transformer as a backbone—reinforced by the use of skip connections from early to deeper layers—was able to capture local and global context simultaneously. This approach architecturally allows the network to maintain high-resolution spatial features while understanding global semantic relationships between objects. Interestingly, however, although this configuration is theoretically superior, its initial performance did not manage to surpass the baseline x13 model, which was designed to be lighter and more efficient. This suggests that the complexity of a transformer architecture does not always directly correlate with performance improvement if not supported by appropriate input and training configurations.

The researcher then explored the use of a Swin Transformer-based backbone more deeply in the context of BEV representation in the tests in table \ref{tab:perbandingan_test_loss_bev_ekstraksi}, using the \textit{mmsegmentation} library. One significant finding was that the SKGE-Swin (stage 1→4) + LiDAR model managed to surpass the performance of all previous models, including x13. This result indicates that the BEV representation—which is a top-down projection of the vehicle's surroundings—is highly compatible with the way the Swin Transformer works, which is based on shifted-window attention. The BEV representation makes it easier for the transformer to understand spatial structure at high resolution and to handle small or distant objects that are typically difficult for a CNN to capture. This is reinforced by the research findings from \cite{li2022bevformerlearningbirdseyeviewrepresentation} in BEVFormer, which proves the effectiveness of Transformers in BEV representation due to their ability to efficiently manage long-range relations between tokens.

The Swin Transformer model also offers advantages when combining multi-modal information such as RGB camera and LiDAR, as seen in the (stage 1→4) + LiDAR configuration. The model's ability to integrate spatial and semantic information from these two sensors reflects great potential for sensor fusion, which, according to \cite{Chitta2022-wl} in TransFuser, can produce more robust perception in complex and dynamic environments. Therefore, in the context of BEV, the Swin Transformer not only acts as a replacement for CNNs but also as a more flexible and scalable architecture for handling diverse and large input data.

To test for performance consistency, the researcher also tried the same model using a different library, namely \textit{Official PyTorch}. The results actually showed a very significant performance increase (see table \ref{tab:perbandingan_ds_bev_ekstraksi}) compared to the x13 baseline as well as previous models that used mmsegmentation. This difference in results indicates that the final performance of a model is not only determined by its architecture but also by implementation details—from how weights are initialized, the preprocessing pipeline, to inference strategies like the use of mixed precision or batchnorm sync. A study by \cite{SHMUELI20164552} showed that framework-level optimization differences can cause performance deviations of up to 10-20\% on segmentation and detection tasks, even when using identical architectures.

Although CNNs remained efficient and superior under strict edge-device or real-time inference conditions, the Swin Transformer architecture proved to be superior in the BEV feature extraction task, especially when supported by multi-stage and multi-modal input configurations. However, this performance is highly sensitive to the implementation framework, which underscores the importance of cross-library validation when evaluating deep learning models for real-world systems.
\begin{figure*}[!t]
    \centering
    \includegraphics[width=1\textwidth]{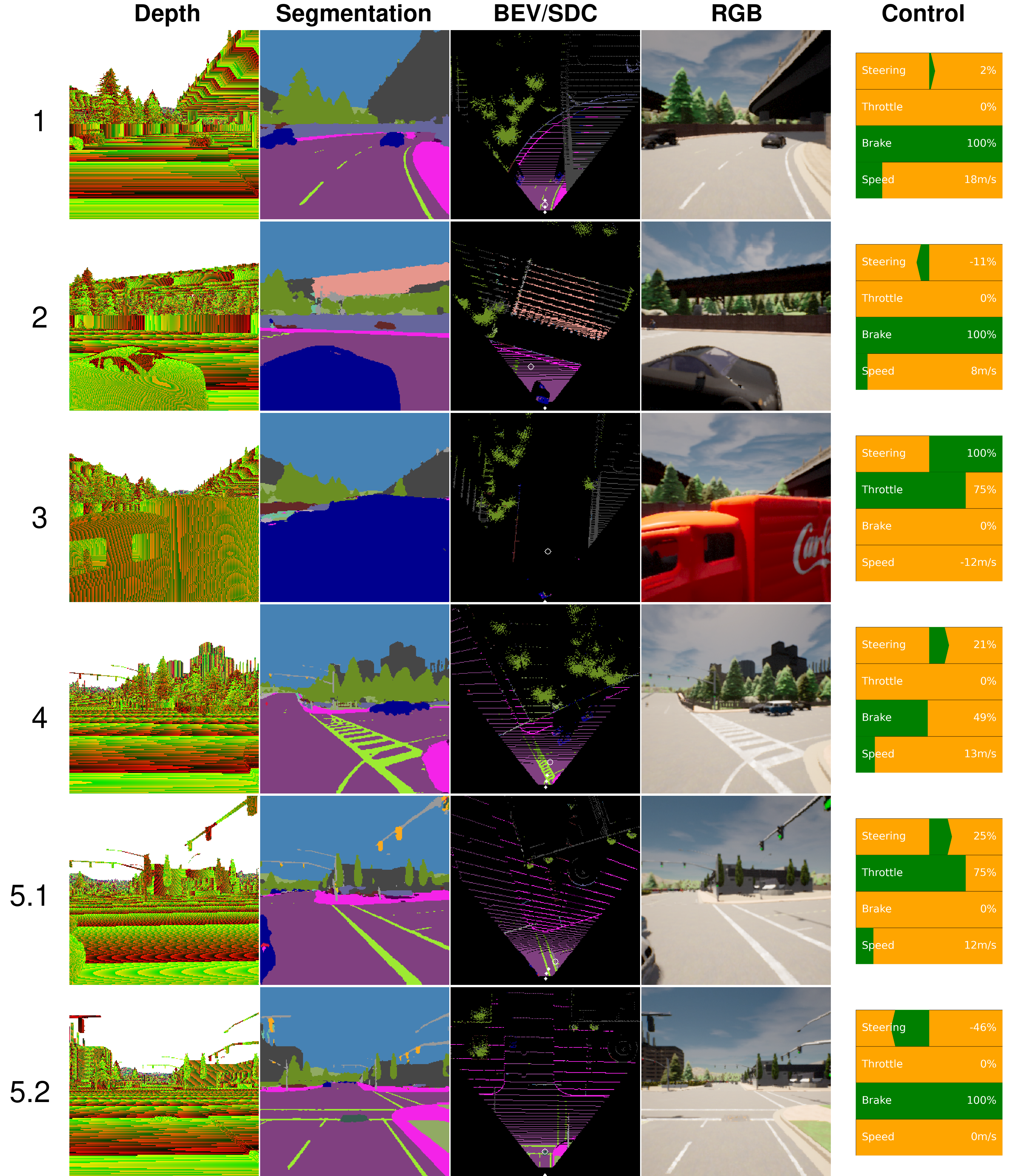}
    \caption{Qualitative Result}
    \label{fig:qualitative_result}
    % \vspace{2mm}
    
    \raggedright
    {\footnotesize
    1. During a turn, the model is able to anticipate well, by applying the brake and reducing the vehicle's speed. \\
    2. A car that suddenly appears from the edge of the image, and is only detected after being very close to the vehicle. The model is able to anticipate well, by braking suddenly. \\
    3. The model is unable to anticipate well (the brake is not applied and the throttle is still on), so the vehicle collides with that vehicle. \\
    4. The vehicle goes off the predetermined path, and is outside the lane. This is possibly because the semantic segmentation considers the vehicle to be on the correct path. \\
    5.1. Using the x13 model. When turning right at an intersection, the model is unable to anticipate well, because it does not look to the left first. \\
    5.2. Using the SKGE-Swin-base (stage 1→4) model. When turning right at an intersection, the model is able to anticipate well, by looking to the left first to ensure no vehicles are approaching from the left. \\
    }
    
    \end{figure*}

\begin{table}[!t]
    \centering
    \caption{Driving Score, Route Completion, and Penalty for Semantic Segmentation Backbone Model (Encoder A)}
    \label{tab:perbandingan_ds_backbone}
    \begin{tabular}{llccc}
    \toprule
    \textbf{Model} & \textbf{RC↑} & \textbf{IP↑} & \textbf{DS↑} \\
    \midrule
    SKGE-Swin-base (stage 4→1) & 65.9631 & 0.3428 & 22.4566 \\
    SKGE-Swin-base (stage 1→4) & 63.2475 & 0.3644 & 22.9775 \\
    SKGE-Swin-base (stage 1,2,3→4) & 71.6942 & 0.2794 & 20.1073 \\
    SKGE-Swin-base (stage 3) & 67.2861 & 0.3495 & 23.4526 \\
    SKGE-Swin-base (stage 2→3) & 68.9117 & 0.3604 & \textbf{24.6327} \\
    \bottomrule
    \end{tabular}
    \end{table}

\begin{table}[!t]
    \centering
    \caption{Driving Score, Route Completion, dan Penalty Model BEV Extraction Feature (Encoder B)}
    \label{tab:perbandingan_ds_bev_ekstraksi}
    \begin{tabular}{cp{3cm}ccc}
    \toprule
    \textbf{Library} &\textbf{Model} & \textbf{RC↑} & \textbf{IP↑} & \textbf{DS↑} \\
    \midrule
    M*&SKGE-Swin-base (stage 3) & 78.7883 & 0.3077 & 24.2636 \\
    &SKGE-Swin-base (stage 2→4) & 72.8196 & 0.3091 & 22.1892 \\
    &SKGE-Swin-base (stage 1→4) & 85.9487 & 0.2703 & 23.2669 \\
    &SKGE-Swin-base (stage 1→4) + lidar & 74.2968 & 0.4245 & \textbf{31.5026} \\
    &SKGE-Swin-tiny (stage 4) & 76.0050 & 0.3673 & 27.9053 \\
    \midrule
    OP\textsuperscript{\textdagger}&SKGE-Swin-tiny (stage 4) & 79.2671 & 0.4106 & 32.7298 \\
    &SKGE-Swin-tiny (stage 1→4) & 82.8075 & 0.4491 & \textbf{37.1046} \\
    &SKGE-Swin-tiny (stage 4) float16 & 76.7540 & 0.3056 & 23.5084 \\
    &SKGE-Swin-tiny (stage 1→4) float16 & 62.7036 & 0.5040 & 31.5970 \\
    \bottomrule
    \end{tabular}
    \vspace{2mm}
    
    \raggedright
    {\footnotesize
    * M: mmsegmentation implementation. \\
    \textsuperscript{\textdagger} OP: Official Pytorch implementation.
    }
\end{table}

Table \ref{tab:perbandingan_test_loss_backbone} shows a performance comparison between backbone models implemented in the mmsegmentation library. From these results, the SKGE-Swin model with a feature extraction sequence from stage 1 to 4 shows the best performance. This indicates that a tiered feature extraction process from beginning to end can capture a more complete spatial representation compared to focusing on just a single stage.
\begin{table}[!t]
    \centering
    \caption{Test Loss between Models with Semantic Segmentation Backbone (Encoder A)}
    \label{tab:perbandingan_test_loss_backbone}
    \begin{tabular}{lccc}
    \toprule 
    \textbf{Model Name} & \textbf{Test Metric↓} \\
    \midrule
    SKGE-Swin (stage 4→1) & 0.9138 \\
    SKGE-Swin (stage 1→4) & \textbf{0.5017} \\
    SKGE-Swin (stage 1,2,3→4) & 0.5774 \\
    SKGE-Swin (stage 3) & 0.5047 \\
    SKGE-Swin (stage 2→3) & 0.5620 \\
    \bottomrule
    \end{tabular}
\end{table}

\begin{table}[!t]
    \centering
    \caption{Test Loss between Models with BEV Feature Extraction (Encoder B)}
    \label{tab:perbandingan_test_loss_bev_ekstraksi}
    \begin{tabular}{clccc}
        \toprule 
        \textbf{Library} &\textbf{Model Name} & \textbf{Test Metric↓} \\
        \midrule
        M*&SKGE-Swin (stage 3) & 0.4668 \\
        &SKGE-Swin (stage 2→4) & 0.4683 \\
        &SKGE-Swin (stage 1→4) & 0.4690 \\
        &SKGE-Swin (stage 1→4) + lidar & \textbf{0.4404} \\
        &SKGE-Swin-tiny (stage 4) & 0.4702 \\
        \midrule
        OP\textsuperscript{\textdagger}&SKGE-Swin-tiny (stage 4) & \textbf{0.4287} \\
        &SKGE-Swin-tiny (stage 1→4) & 0.4325 \\
        &SKGE-Swin-tiny (stage 4) float16 & 0.5468 \\
        &SKGE-Swin-tiny (stage 1→4) float16 & 0.9682 \\
        \bottomrule
        \end{tabular}
        \vspace{2mm}

    \raggedright
    {\footnotesize
    * M: mmsegmentation implementation. \\
\textsuperscript{\textdagger} OP: Official PyTorch implementation.
    }
\end{table}

\subsection{Model Behaviour}

In this section, the researcher presents qualitative results from the trained model, focusing on how the model behaved in various driving situations. The following images show how the model interpreted its surroundings, as well as its response to various conditions such as turns, the presence of pedestrians, and traffic. In this section, the researcher used the SKGE-Swin-tiny (stage 1→4) model, which was the model with the highest driving score in the previous experiments. This model had been trained with various data and was able to provide a good response to different conditions.

In figure \ref{fig:qualitative_result}.1, it can be seen that the best model used by the researcher responsively handled dynamically changing road conditions, especially when the vehicle entered a sharp turn. The model's decision to apply the brakes was an attempt to reduce oversteering due to high speed, and this could be attributed to the model's ability to understand the semantic structure of the road from the segmentation results and the Bird's Eye View (BEV) representation. Previous research by \cite{li2022bevformerlearningbirdseyeviewrepresentation} on BEVFormer showed that BEV representation can help in strengthening spatial perception and dynamic movement prediction, including road contours and drivable areas. This supports the assumption that the BEV representation in the model provides an advantage in contextually and anticipatorily responding to turns.

The model also showed adaptive behavior in dealing with the presence of pedestrians and vehicles that appeared suddenly in figure \ref{fig:qualitative_result}.2, where a sudden braking response was performed. This behavior indicates that the model is able to recognize dynamic objects and process safety priorities correctly. This result is in line with the study from \cite{philion2020liftsplatshootencoding} in Lift-Splat-Shoot, which showed that BEV-based segmentation can help in responding to fast-moving objects, as it can efficiently combine temporal-spatial information. However, this ability is highly dependent on the resolution and quality of the segmentation prediction, which is still a challenge when objects are partially occluded or appear from a lateral direction.

Conversely, in figure \ref{fig:qualitative_result}.4, a weakness of the model in identifying off-route conditions was found. Although semantically the vehicle was considered to still be on the correct path, its actual position had already deviated from the specified route. This indicates that semantic segmentation has limitations in understanding macro-level trajectory context or road boundaries that are not always clearly visible. This finding is supported by \cite{caesar2020nuscenesmultimodaldatasetautonomous} in the nuScenes benchmark, which states that excessive reliance on semantics can lead to misperception of context when visual sensor data is degraded or lacks information. To overcome this, cross-validation between HD maps (high-definition maps), odometry, and semantic segmentation results is necessary.

In figure \ref{fig:qualitative_result}.3, the model's failure to anticipate a vehicle appearing suddenly from a lateral direction, leading to a collision, was shown. This weakness reflects the model's limitations in understanding lateral dynamics and the very narrow response time. A study by \cite{li2022bevdepthacquisitionreliabledepth} regarding BEVDepth stated that accuracy in lateral detection still depended on the camera's point of view and the feature aggregation time from previous frames. This suggests that the ideal solution may involve a combination of 360° surround cameras, radar, or multi-modal sensor fusion integration to increase awareness of objects from the side.

Finally, the comparison between the SKGE-Swin-tiny (stage 1→4) and x13 models in figure \ref{fig:qualitative_result}.5 shows the superiority of the transformer in understanding the situational context at road intersections. The ability of SKGE-Swin-base to "look to the left" before turning right indicated that this model had better situational awareness; for more details see figure \ref{fig:architecture-bev-tabrakan}. This is strongly suspected to originate from the skip stage connection implemented at the Swin Transformer, which allows the model to maintain high-resolution spatial information while also understanding global context. This is in line with the findings from \cite{He2015-lm} in ResNet, which states that skip connections can help the model learn more complex features without losing important spatial information. In contrast, the x13 model, which is based on CNN, tends to focus more on local features and may not be able to capture broader contextual information as effectively as the Swin Transformer. This is supported by \cite{Liu2021-qz}, which explains that the hierarchical and shifted window architecture in Swin explicitly supports the preservation of local and global context simultaneously. Therefore, these results showed that the architectural design had a major impact on the quality of the model's decision-making in the context of complex navigation.

% . The x13 model, which is based on CNN, tends to focus more on local features and may not be able to capture broader contextual information as effectively as the Swin Transformer.
% at Swin Transformer which preserves important spatial information at high resolution. \cite{Liu2021-qz} in Swin Transformer explains that the hierarchical + shifted window architecture in Swin explicitly supports the preservation of local and global context simultaneously. Therefore, these results showed that the architectural design had a major impact on the quality of the model's decision-making in the context of complex navigation.
\begin{table*}[!t]
    \centering
    \caption{Detailed Test Metrics per Component for Each Model (Encoder A)}
    \label{tab:detail_loss_backbone_semantic}
    \resizebox{\textwidth}{!}{%
    \begin{tabular}{p{4cm}cccccccc}
    \toprule
    \textbf{Model Name} & \multicolumn{1}{l}{\textbf{ss\_metric↑}} & \multicolumn{1}{l}{\textbf{wp\_metric↓}} & \multicolumn{1}{l}{\textbf{str\_metric↓}} & \multicolumn{1}{l}{\textbf{thr\_metric↓}} & \multicolumn{1}{l}{\textbf{brk\_metric↓}} & \multicolumn{1}{l}{\textbf{redl\_metric↑}} & \multicolumn{1}{l}{\textbf{stops\_metric↑}} \\
    \midrule
    x13 & 0.8869 & 0.1504 & 0.0182 & 0.0635 & 0.0621 & 0.9705 & 0.9942 \\
    \hline
SKGE-Swin (stage 4→1) & 0.8907 & 0.1799 & 0.0257 & 0.0665 & 0.4064 & 0.8794 & 0.9942 \\
SKGE-Swin (stage 1→4) & 0.8898 & 0.1788 & 0.0262 & 0.0753 & 0.0723 & 0.9665 & 0.9942 \\
SKGE-Swin (stage 1,2,3→4) & 0.8934 & 0.1735 & 0.0222 & 0.0744 & 0.0746 & 0.8794 & 0.9942 \\
SKGE-Swin (stage 3) & 0.8874 & 0.1750 & 0.0233 & 0.0732 & 0.0727 & 0.9576 & 0.9942 \\
SKGE-Swin (stage 2→3) & 0.8871 & 0.1679 & 0.0233 & 0.0649 & 0.0669 & 0.8794 & 0.9942 \\
% SKGE-Swin (stage 2→4) & 0.8811 & 0.1627 & 0.0194 & 0.0690 & 0.0637 & 0.9708 & 0.9942 \\
    \bottomrule
\end{tabular}%
}
\end{table*}

\section{Conclusion}
This research aims to improve the performance of end-to-end driving planning models by utilizing the Swin Transformer as a backbone for feature extraction. The experimental results show that the SKGE-Swin (stage 1→4) model succeeded in achieving the highest driving score of 37.10, which significantly outperforms the other models. This improvement can be attributed to the Swin Transformer's ability in capturing spatial and local representations, complemented by the skip stage which plays a very important role in referencing features from the shallow layer to the deep layer. Evaluation uses the driving score metric as well as qualitative analysis of the model's behavior in various road scenarios. Meanwhile, a baseline model such as x13 shows poor performance, especially in complex decision-making scenarios.

Qualitative evaluation supports these quantitative results. In real scenarios such as the presence of pedestrians, sharp turns, or vehicles coming from the side, the SKGE-Swin model shows wiser and more adaptive responses. The model is able to anticipate the presence of objects earlier and provide actions that are in line with the context, such as slowing down or stepping on the brakes. Conversely, models with lower scores tend to show reckless behavior, such as failing to see the direction of traffic before turning or not braking when needed, which leads to accidents. However, this evaluation also highlights that even the best model still has weaknesses, especially in handling objects that suddenly appear from the side of the road or in conditions of erroneous semantic segmentation. This shows that although a high driving score can be a good performance indicator, a comprehensive assessment must still involve visual analysis of complex scenarios.

For future development, some suggestions that can be given are the improvement of lateral object detection: The model needs to be trained more focusedly on the detection and anticipation of objects coming from the side, such as vehicles from the right or left, which proved to be one ofthe failure points in the qualitative experiments. This can be improved by adding data augmentation from lateral angles or by including side cameras. Addition of Ensemble Methods: To produce a more robust model overall, object detection for traffic lights can be added so that the model can be better at anticipating red or green lights.

\bibliography{reference}

\end{document}